%% file: main.tex
\documentclass[runningheads]{llncs}

\usepackage{eccv}

\usepackage{eccvabbrv}

\usepackage{graphicx}
\usepackage{booktabs}
\usepackage{amsmath}
\usepackage{amssymb}
\usepackage{pifont}

\usepackage[accsupp]{axessibility}  

\usepackage{hyperref}

\begin{document}

\title{REST–HANDS: Rehabilitation with Egocentric Vision Using Smartglasses for Treatment of Hands after Surviving Stroke
} 

\titlerunning{REST-HANDS}

\author{Wiktor Mucha\inst{1} \and
Kentaro Tanaka\inst{2} \and
Martin Kampel\inst{1}}

\authorrunning{W.~Mucha et al.}

\institute{Computer Vision Lab, TU Wien, Favoritenstr. 9/193-1, 1040 Vienna, Austria \email{\{wiktor.mucha,martin.kampel\}@tuwien.ac.at} \and
Trinity Centre for Practice and Healthcare Innovation, School of Nursing and Midwifery, Trinity College Dublin, Dublin, Ireland\\
\email{ktanaka@tcd.ie}}

\maketitle

\begin{abstract}
 Stroke represents the third cause of death and disability worldwide, and is recognised as a significant global health problem. A major challenge for stroke survivors is persistent hand dysfunction, which severely affects the ability to perform daily activities and the overall quality of life. In order to regain their functional hand ability, stroke survivors need rehabilitation therapy. However, traditional rehabilitation requires continuous medical support, creating dependency on an overburdened healthcare system. In this paper, we explore the use of egocentric recordings from commercially available smart glasses, specifically RayBan Stories, for remote hand rehabilitation. Our approach includes offline experiments to evaluate the potential of smart glasses for automatic exercise recognition, exercise form evaluation and repetition counting. We present REST-HANDS, the first dataset of egocentric hand exercise videos. Using state-of-the-art methods, we establish benchmarks with high accuracy rates for exercise recognition (98.55\%), form evaluation (86.98\%), and repetition counting (mean absolute error of 1.33). Our study demonstrates the feasibility of using egocentric video from smart glasses for remote rehabilitation, paving the way for further research. The dataset and code are under 
  \url{https://github.com/wiktormucha/rest-hands}.
\end{abstract}

\section{Introduction}
\label{sec:intro}

\input{sections/1_introduction}

\section{Related Work}
\label{sec:related_work}

\input{sections/2_related_work}

\section{REST-HANDS}
\label{sec:dataset}

\input{sections/3_rest-hands}

\section{Evaluation}
\label{sec:evaluation}

\input{sections/4_experiments}

\section{Conclusion}

\label{sec:conclusion}
\input{sections/5_conclusion}

\subsubsection{Acknowledgements} 

This research was supported by VisuAAL ITN H2020 (grant agreement no. 861091) and the Austrian Research Promotion Agency (grant agreement no. 49450173).

\bibliographystyle{splncs04}
\bibliography{bibliogaphy}
\end{document}

%% file: sections/1_introduction.tex
One of the most significant global health problems is stroke (also known as cerebrovascular accident). It is a medical condition of cerebral blood flow that leads to a sudden loss of brain function, often with severe consequences for those affected. Despite advances in medical care and rehabilitation, stroke remains the third leading cause of mortality and disability worldwide, as reported in the collaborative study by Feigin et al. \cite{feigin2021global}. Hand function is one of the most pressing and persistent challenges for stroke survivors. Studies suggest that approximately 85\% of stroke patients worldwide experience hand dysfunction, with a significant proportion, nearly 60\%, continuing to struggle with upper limb problems, particularly in the fingers and wrists, even after treatment and discharge \cite{cao2021efficacy}. This debilitating hand dysfunction has a significant impact on a person's ability to perform basic Activities of Daily Living (ADLs) that involve grasping and manipulating objects. As a result, it profoundly affects the overall quality of life of stroke survivors \cite{shi2021effects}. To meet this challenge, patients require a rehabilitation process, often involving medical personnel~\cite{cao2021efficacy}, which creates a dependency on the healthcare system. The current trend of an ageing population leads to an increasing demand for medical staff that cannot be met and needs to be addressed in other ways, e.g. assistive technologies are proposed in the literature \cite{ballester2024vision,text2taste, adeli2024benchmarking}.

Driven by this need, we propose to develop a remote hand rehabilitation tool for stroke patients using egocentric video captured by commercially available smartglasses, such as RayBan Stories. 
Current solutions for remote rehabilitation rely on the user to perform exercises in front of a vision system, which demands additional setup \cite{nikishina2019application} and lack automation as they require analysis by healthcare professionals and use expensive Virtual Reality (VR) equipment~\cite{tada2022quantifying,herrera2023rehab,tada2024integrated}. Smartglasses, on the other hand, are tools with greater potential to perform multiple tasks in the future~\cite{plizzari2024outlook}, such as action recognition \cite{effhandegonet} and assistance \cite{text2taste}, one of which could be assistance in rehabilitation. As RayBan Stories only allows for video recording without any processing on the device or real-time data transfer, we intend to show the potential of this technology for remote rehabilitation tasks through offline experiments to encourage research and development in this direction. This type of rehabilitation tool requires several automatic functionalities, including exercise recognition, exercise form evaluation, and repetition counting to track progress. Our study is, to the best of our knowledge, the first attempt to use egocentric vision for automatic remote post-stroke hand rehabilitation. To experimentally evaluate this approach and answer the question of whether smartglasses with camera are feasible for remote hand rehabilitation, we create a REST-HANDS dataset containing labels for these three tasks and establish benchmarks for each. Fig.~\ref{fig:rehab} shows an overview of our approach. Our contributions are the following:

\begin{figure}[t]
  \centering
  \includegraphics[width=1\linewidth]{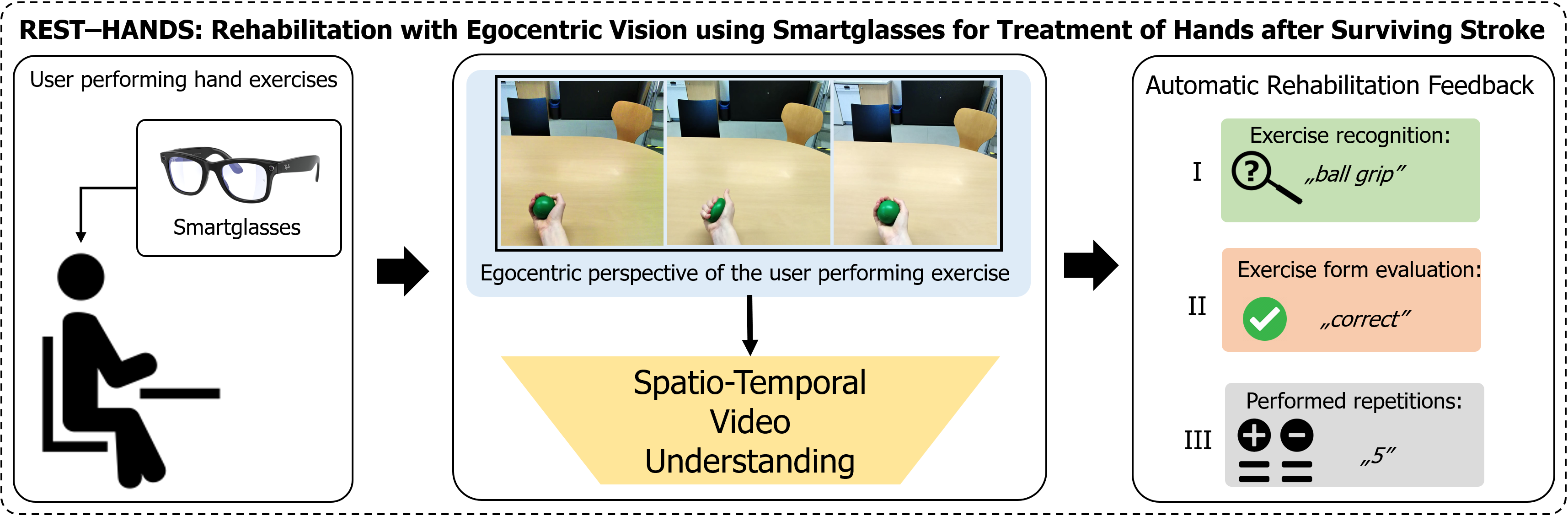}
  \caption{We propose REST-HANDS. Automatic remote hand rehabilitation assessment system for stroke survivors at home using egocentric vision through smartglasses. The images captured from the user's perspective are used to analyse the rehabilitation process with a spatio-temporal deep learning network to recognise exercises, evaluate exercise form and count repetitions. To assess the proposed idea, we create a first egocentric dataset with hand exercises for stroke patients and establish baseline results for three tasks by implementing and comparing different state-of-the-art architectures. }
  \label{fig:rehab}

\end{figure}

\begin{itemize}
   \item We propose to use egocentric videos captured by comfortable smartglasses for automatic assessment of remote hand rehabilitation for stroke survivors. 
    
    \item We create the first dataset of hand exercises for stroke patients from an egocentric perspective using smartglasses. The dataset consists of videos of exercises with the following labels: exercise class, evaluation form assessed by an expert physiotherapist, and number of repetitions per exercise. 
    
    \item We employ state-of-the-art video understanding methods to implement solutions for the most required tasks in automated remote rehabilitation: exercise recognition, exercise form evaluation and repetition counting.

    \item We present an extensive evaluation, ablations and experiments on the introduced dataset to prove the feasibility of our approach, justify implementation details and discover key areas for future research. Our benchmarks in equally distributed subsets result in exercise recognition with an accuracy of 98.55\%, 86.98\% accuracy of form evaluation and repetition count with Mean Absolut Error equal to 1.33.

\end{itemize}

The paper is structured as follows: Section \ref{sec:related_work} presents related work on remote hand rehabilitation and monitoring systems, together with recent methods for spatio-temporal video understanding, highlighting areas for improvement. Section \ref{sec:dataset} describes our dataset for the egocentric understanding of hand rehabilitation, called REST-HANDS, together with the methods we propose to solve rehabilitation-related tasks. Evaluation and experiments are presented in Section~\ref{sec:evaluation}. Section \ref{sec:conclusion} concludes the study with its main findings.

%% file: sections/2_related_work.tex


\paragraph{\textbf{Remote Vision-based Hand Monitoring and Rehabilitation.}}

Several studies propose automated computer vision systems to monitor hand functionality for people who have survived a stroke. Some of the studies dedicated to hand monitoring and rehabilitation, apart from focusing on stroke, are aimed at other patients, for example, people after cervical Spinal Cord Injury (cSCI). Likitlersuang et al.~\cite{likitlersuang2019egocentric} implement an egocentric hand monitoring system for cSCI patients to assess hand-object interactions. Captured egocentric video is processed to detect the hand, segment the hand outline, and distinguish the user's left or right hand. This information is employed to perform binary detection of functional interactions of the hand with objects while performing ADLs. This interaction detection allows analysis of hand use in the home, such as the number of interactions per hour, the duration of interactions and the percentage of interactions over time. Visee et al.~\cite{visee2020effective} propose an egocentric hand detection method to create a hand-tracking tool for monitoring cSCI patients. The experiments are carried out with real cSCI patients, but the study is restricted to hand detection. Dusty and Zariffa~\cite{dousty2020tenodesis} propose a method to enable clinicians and researchers to monitor the use of the tenodesis grasp by people with cSCI at home, leading to remote therapeutic guidance. Their approach includes hand detection, pose estimation and arm orientation estimation to assess wrist angle data leading to the detection of the tenodesis grasp.


A separate group of works focuses on the use of VR devices to create interactive systems for a user in the form of game-like tasks. Tada et al.~\cite{tada2022quantifying} desgines rehabilitation drills in VR using Microsoft's HoloLens2. A similar system is created by Herrera et al.~\cite{herrera2023rehab}, who also provide a framework for creating VR game drills dedicated to rehabilitation. In an extension of previous work, Tada et al.~\cite{tada2024integrated} extend VR to Mixed Reality (MR) with the possibility of interacting with dynamic objects. All of these proposed systems aim to monitor cognitive and motor functions for further expertise evaluation during task performance.


In contrast to studies that focus on the creation of VR rehabilitation tasks, there are studies that address the monitoring of exercises and the measurement of performance. Nikishina et al.~\cite{nikishina2019application} propose an exocentric video-based system for exercise recognition. It consists of a camera placed frontally, and the user repeats the exercise displayed on the computer screen. Recognition is performed with an SVM-based algorithm.
Placidi et al.~\cite{placidi2023patient} present a framework to facilitate collaboration between patients and therapists in the context of rehabilitation. It allows real-time hand tracking at 40 FPS. The key innovation of this work is its ability to provide therapists with quantitative data on the mobility of each hand joint, in addition to qualitative assessments, allowing detailed monitoring of rehabilitation progress. Our study differs by focusing on automated exercise tracking to reduce the engagement of medical staff in remote rehabilitation.

\paragraph{\textbf{Video Understanding.}}

Initially, traditional 2D Convolutional Neural Networks (CNNs) were extended to video by processing individual frames independently. However, this approach struggles to effectively capture temporal dynamics. This limitation led to the development of 3D CNNs, such as the C3D model, which uses 3D convolutions to extract spatio-temporal features from video clips~\cite{tran2014learning} by extending the kernel into the temporal dimension. Following C3D, many 3D CNN architectures have been proposed, such as I3D~\cite{carreira2017quo}, S3D-G~\cite{xie2018rethinking} or SlowFast~\cite{feichtenhofer2019slowfast}. Another approach to capturing temporal information are two-stream networks, which use optical flow information alongside CNNs that embed spatial features~\cite{simonyan2014two}. Donahue et al~\cite{donahue2015long} propose the use of Long Short-Term Memory (LSTM) networks to extract temporal information. LSTMs are designed to handle sequential data and capture long-term dependencies, enabling the detection of activity in videos by modelling the temporal dynamics of frame features. More recently, transformers, originally developed for natural language processing tasks, have been adapted for video understanding due to their ability to model long-range dependencies and capture contextual information. The Vision Transformer (ViT)~\cite{dosovitskiy2020image} processes video frames as sequences of patches and applies self-attention mechanisms to learn spatio-temporal features. Recent models extend this approach by incorporating temporal attention mechanisms~\cite{bertasius2021space} or by using 3D patches~\cite{li2022mvitv2,liu2022video}.

In our study, we perform the tasks of (1) exercise recognition, addressed as fine-grained video classification, and (2) exercise form evaluation, similar to action quality assessment. We evaluate the exercise form between correct and incorrect repetitions through binary classification. The methods described above have been shown to successfully perform video classification~\cite{feichtenhofer2019slowfast,li2022mvitv2,liu2022video,carreira2017quo} and action quality assessment~\cite{parmar2019and, zhou2022uncertainty, zhang2024auto} due to their ability to extract spatio-temporal features. Our third task is (3) repetition counting. Early research relies on visual content alone, compressing the motion field of video into one-dimensional signals and counting repetitive activity using methods such as Fourier analysis~\cite{azy2008segmentation} or peak detection~\cite{thangali2005periodic}. However, spatial and spatio-temporal feature extractors are the key component for repetition counting as well in more recent studies~\cite{zhang2020context,dwibedi2020counting}.

\paragraph{\textbf{In contrast to previous studies}} that have focused on hand rehabilitation using VR headsets or exocentric cameras, our work employs user-friendly smartglasses that capture the person's perspective. In addition to comfort, the advantage of smartglasses is that as research in egocentric vision progresses, these devices are likely to be widely used in our lives and serve multiple purposes~\cite{plizzari2024outlook}. To date, research in hand rehabilitation has only used egocentric perspective images to perform basic tasks related to hand tracking and monitoring~\cite{visee2020effective,likitlersuang2019egocentric,dousty2020tenodesis}. More detailed information about hand motor functionalities is provided by the use of VR or MR headset tools~\cite{tada2022quantifying,herrera2023rehab,tada2024integrated} but without the capability of automatic analysis.
Our study stands out as a step towards automated remote rehabilitation. To the best of our knowledge, we create the first dataset of hand rehabilitation exercises recorded with smartglasses. In contrast to~\cite{visee2020effective,likitlersuang2019egocentric,dousty2020tenodesis,placidi2023patient} and VR/MR~\cite{tada2022quantifying,herrera2023rehab,tada2024integrated}, which lack automatic analysis tools and provide hand information that needs to be analysed by a healthcare professional, we focus on higher inference tasks including exercise recognition, form evaluation and repetition counting using state-of-the-art approaches for video understanding.

%% file: sections/3_rest-hands.tex
Our study proposes an approach for remote hand rehabilitation for stroke survivors using egocentric vision from smartglasses. In order to provide automated assistance, progress tracking or support, such a system must at least be able to recognise exercises, determine whether the user is doing them correctly or not, and track the number of repetitions. At the time of this study, no egocentric dataset exists for this purpose. Therefore, we create a new dataset called REST-HANDS. Furthermore, we provide baseline results for exercise recognition, exercise form evaluation and repetition counting using state-of-the-art methods.

\subsection{REST-HANDS Dataset}

\begin{figure}[t]
  \centering
  \includegraphics[width=1\linewidth]{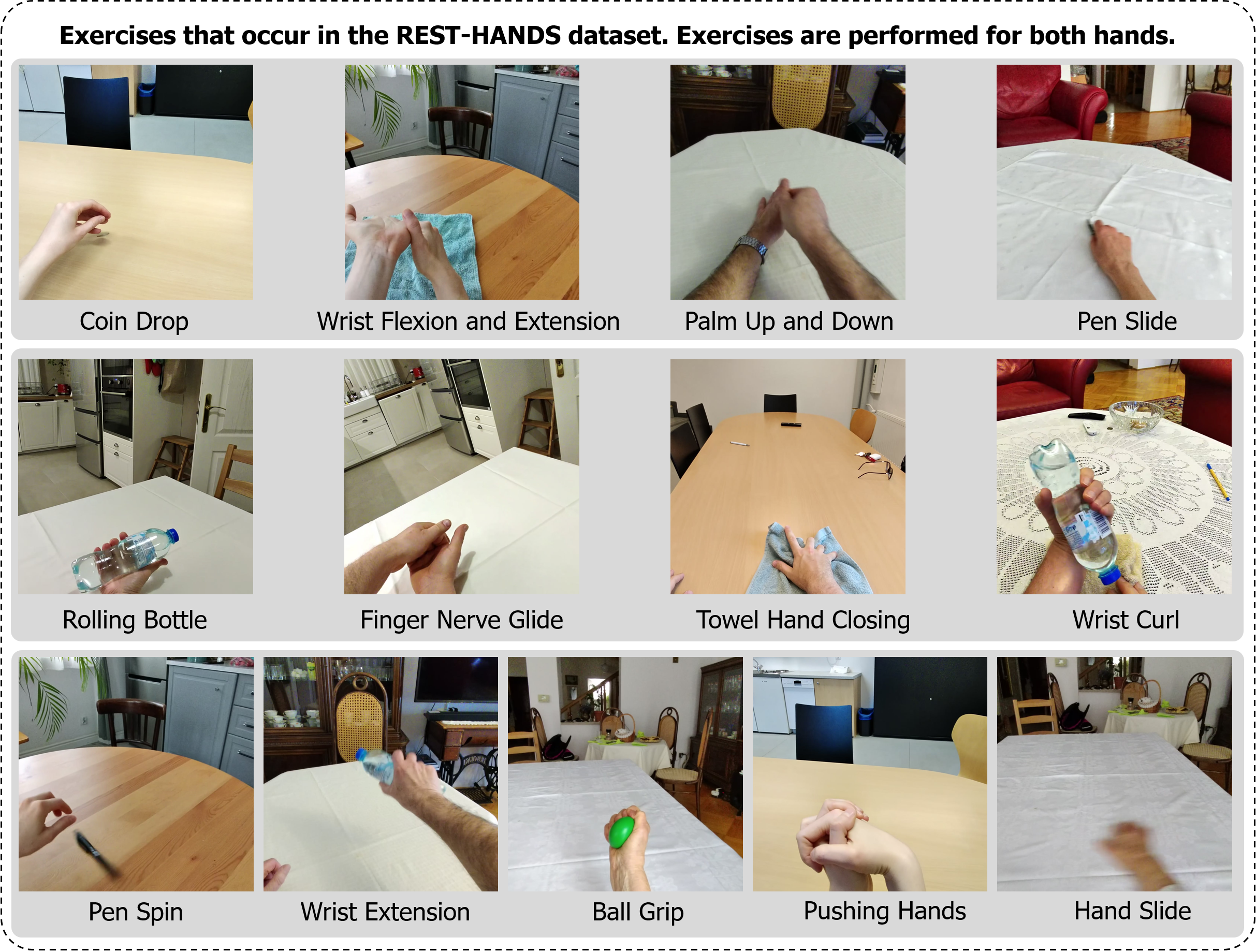}
  \caption{Examples of frames from REST-HANDS dataset representing different exercises included in the dataset. Each exercise except \textit{Pushing Hands}, which involves both hands, is performed and categorised separately for each hand resulting in 25 classes.
  }
  \label{fig:dataset}

\end{figure}

\begin{figure}[t]
  \centering
  \includegraphics[width=1\linewidth]{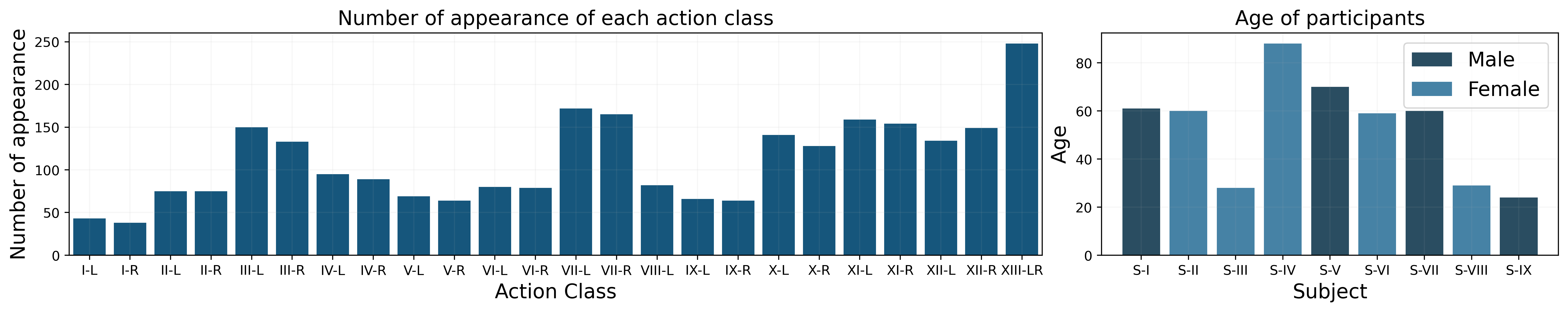}
  \caption{Distribution of exercise classes and age of participants in REST-HANDS. }
  \label{fig:stats}

\end{figure}

We present the REST-HANDS dataset - the first egocentric hand rehabilitation dataset from smartglasses recordings - with the aim of raising the scientific community's interest in hand rehabilitation using egocentric vision. The dataset contains twelve exercises repeated for each hand and one exercise involving both hands, resulting in a total of twenty-five exercise labels. The choice of exercises is made in consultation with a health professional specialising in physiotherapy, who, in addition, creates labels for exercise form evaluation by analysing recordings of participants. Finally, the dataset includes labels for counting the number of repetitions performed by the user in the range from 1 to 20. REST-HANDS contains 197.8K frames representing hand exercises, making it larger than other egocentric hand-based action understanding datasets such as H2O~\cite{Kwon_2021_ICCV} with 90.7K frames and FPHA~\cite{garcia2018first} with 105K frames. Examples of recordings of users performing exercises that are part of REST-HANDS are shown in Fig. \ref{fig:dataset}.

\subsubsection{Data collection.}

The videos are recorded by 9 participants, each wearing RayBan Stories smartglasses, which capture a first-person perspective in 1264 $\times$ 1264 pixels and 30 Hz. This device offers a high level of comfort to the user and, with ongoing research into egocentric vision, has the potential to be a versatile tool for various tasks beyond rehabilitation \cite{plizzari2024outlook}. The advantage of egocentric approaches as opposed to exocentric ones is that the user is not tied to a location and can perform the exercises wherever they wish. Participants include 5 women and 4 men aged between 24 and 88 years, with one being a stroke survivor. The broad age range brings diversity in data and allows for the assessment of our approach across various demographic groups. The recordings are made in home scenarios to mimic the real-world environment of remote rehabilitation, increasing the practical applicability of the findings to home therapy. All participants are volunteers who understand the purpose of the study and give their informed consent. The exercises are first explained to the participants and then performed for 30 seconds. Participants are not corrected during the exercises, except for critical mistakes, in order to provide real-life conditions and introduce variability into the data. Fig.~\ref{fig:stats} presents the distribution of the exercise classes and the data of the participants. It is unbalanced as participants perform exercise for a given time, and each exercise varies in difficulty or pace, leading to a different number of registered instances.

\subsubsection{Data labelling process.}The dataset is manually annotated to ensure accurate annotations, with the help of a health professional, which is critical for subsequent analysis and to ensure the application of our approach in clinical and research settings. The labelling process includes several key steps.

Exercise type and duration are manually labelled. This initial labelling process involves watching the video and categorising the exercises based on their specific movements. In order to provide accurate information on the performance of the exercises, each video clip is carefully reviewed and timestamps are added to indicate the start and end of each movement. This detailed annotation allows individual exercise segments to be extracted from the continuous video recordings. The timestamping process ensures that each repetition of exercise is clearly defined, facilitating accurate analysis of movement duration and transitions. In total, REST-HANDS contains 2.7K clips with exercises for the recognition tasks.

Exercise correctness annotations are created by a health professional who specialises in physiotherapy. The physiotherapist reviews the video clips and assigns a binary label (correct or incorrect) every 10 seconds based on established clinical criteria. Segments where the form is not clear are analysed at a finer level, separately for each repetition. For repetitions where it is not possible to determine correctness, e.g. due to occlusion, repetitions are discarded. The final number of exercises for form evaluation equals 2.7K.

Repetition counting annotations are created by randomly sampling recordings representing exercises using timestamp information about repetitions. From each video, we randomly sample segments ranging in length from one repetition to the maximum number of repetitions that occur in the video, and no greater than twenty. These segments partially overlap, but each one is unique, which enlarges the data. In total, there are 2.5K clips for repetition counting.

\subsection{Methods}

To address the challenges and establish benchmarks in the REST-HANDS, we propose to use state-of-the-art architectures that are proven to be effective in large-scale video understanding datasets, such as Kinetics-400~\cite{kay2017kinetics} or Epic-Kichens~\cite{Damen2018EPICKITCHENS}, and adapt them to the REST-HANDS tasks. For exercise recognition and evaluation, we process the sequences with a network that extracts spatio-temporal features representing the scene. We build on three state-of-the-art networks, including 3D~CNN and transformer-based architectures. This representation is further processed by the MLP network dedicated to the target task. Repetitions are counted through a pick-detecting CNN and a counter.

\begin{figure}[t]
  \centering
  \includegraphics[width=1\linewidth]{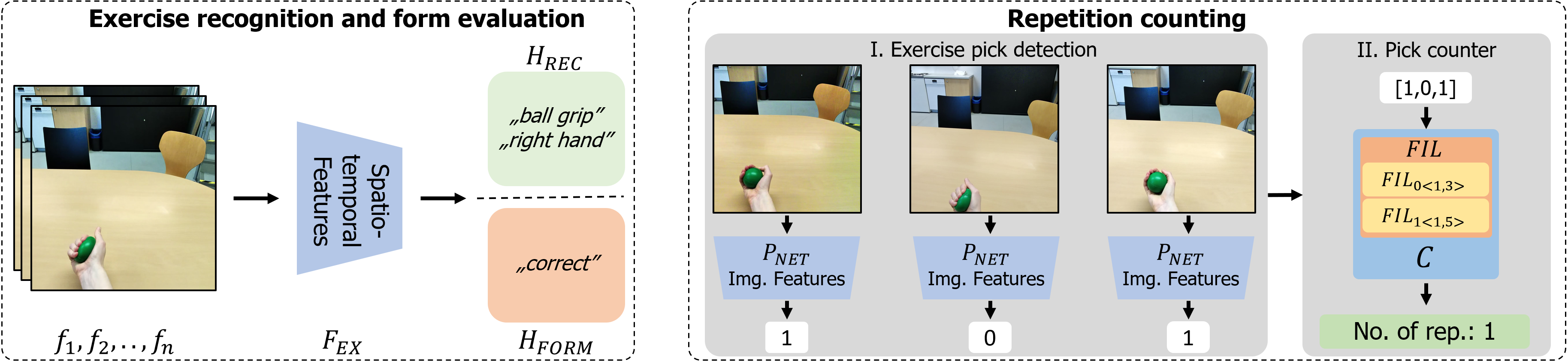}
  \caption{Our methods for exercise recognition, form evaluation and repetition counting. }
  \label{fig:methods}

\end{figure}

\subsubsection{Exercise Recognition and Form Evaluation.}

Input sequence $I_{SEQ}$ representing a user performing hand exercise is represented by frames $f_n,$ where $n \in <0,1,..,N>$. First, the sequence is passed through the feature extraction stage $F_{EX}$, resulting in feature map $F_M$. $F_M$ is processed with prediction head $H$ depending on the task. To establish benchmarks and find the best architecture, the sequences are processed with three distinct $F_{EX}$: a 3D CNN SlowFast~\cite{feichtenhofer2019slowfast}, transformer-based Swin3D~\cite{liu2022video} and MViTv2~\cite{li2022mvitv2}. 


The first $F_{EX}$ implemented, SlowFast \cite{feichtenhofer2019slowfast}, employs a 3D~CNN to take single-stream input and operates on two various framerates, slow and fast. The slow path is built from 3D ResNet50 \cite{he2016deep} with a large temporal stride $\tau$ on input frames. The fast pathway is a similar network which processes input with a stride $\tau/\alpha$, meaning $\alpha$ times denser information representation.

The second $F_{EX}$ is a transformer-based architecture. In the original ViT \cite{dosovitskiy2020image}, input images, $x \in \mathbb{R}^{H\times W\times C}$, are flattened into 2D patches, $x_p \in \mathbb{R}^{N\times(P^2\cdot C)}$, and merged with a learnable token. Standard 1D positional embedding is added to the sequence to preserve positional information. In the next step, the sequence is processed through a transformer encoder, which consists of alternating layers of Multi-head Self-Attention (MSA) and MLP blocks with layer normalisation at the input and residual connection at the output. In our implementation, we follow Swin3D \cite{liu2022video} and modify the original ViT to use 3D patches over the input sequence. The result of linear embedding is passed through a 3D-shifted window-based MSA mechanism. This overcomes the limitation of the original MSA in understanding connections between windows due to a non-overlapping partitioning strategy. For the latter, windows are shifted along the temporal, height and width axes.

For the third $F_{EX}$, we modify the ViT architecture following changes introduced in MViTv2 \cite{li2022mvitv2}. We add relative positional embeddings where the relative position between the two input elements $i$ and $j$ is encoded in the positional embedding $Rp(i),p(j) \in R^d$, where $p(i)$ and $p(j)$ denote the spatio-temporal position of an element $i$ and $j$, respectively. The complexity and memory requirements in attention blocks are reduced with a residual pooling connection added within the attention block by adding the pooled query tensor to the output, resulting in $Z := Attn(Q,K,V) + Q$.


The $F_M$ output of the $F_{EX}$ stage is processed by a prediction head~$H$. Both tasks, exercise recognition and exercise form evaluation, are treated as fine-grained video classification tasks. For exercise recognition, the network $H_{REC}$ is built as an MLP with an output size of 25 for each exercise class. The form evaluation is a binary classification problem with a modification of MLP to 2 output classes resulting in $H_{FORM}$.

\subsubsection{Repetition Counting Based on Exercise Pick Detection.}

The purpose of repetition counting is to determine how many times a pattern is repeated in the input sequence $I_{SEQ}$. To establish a benchmark on our dataset, we propose a method based on the summation of the picks $P$ appearing in $I_{SEQ}$, where $P$ is understood as the beginning of the repetition. The advantage of this method is that the sequence can be processed frame by frame or in short sequences, rather than processing the entire video, which is computationally expensive.  Embedding the whole clip might result in information loss, as the input has to be of a fixed size for each sequence, and the exercises vary in length as they are performed at different paces. Our method uses a CNN network, $P_{NET}$, specifically EfficientNetv2s~\cite{tan2021efficientnetv2}, to extract spatial image features and estimate whether the given frame $F_n$ represents $P$ through a binary classification head, denoted as $H_{P}$. As estimating the exact $P$ of exercise in 30 Hz video is not possible, we consider $P$ to be a sequence of $F_n$, where $n \in <0,1,..,5>$. Each $I_{SEQ}$ is processed to determine $F_n$ whether it is a pick or not, resulting in a binary sequence i.e. $[1,1,0,0]$ meaning the repetition has started. The sequence is processed by a zero crossing counter $C$ which embeds the filter module $FIL$ with a noise spike filter $FIL_0$ and $FIL_1$ to eliminate out values $\in \{0,1\}$ that occur singularly and in short sequences considered as noisy $P$ predictions from the $P_{NET}$.

%% file: sections/4_experiments.tex
We showcase the feasibility of remote hand rehabilitation from egocentric vision and establish benchmark results for three independent tasks in the REST-HANDS dataset required for rehabilitation progress tracking, namely, exercise recognition, exercise form evaluation and repetition counting. We perform a series of experiments and ablations to highlight the research challenges in egocentric remote hand rehabilitation and justify our design choices.

We conduct the evaluation using two different training and testing splits: equal distribution of test classes and Leave-One-Out Cross-Validation (LOOCV), where one subject is selected for testing and the remaining data is split between training and validation in a 90/10 ratio. The reason for including LOOCV is that, although equal distribution testing is standard practice for evaluation, in a real-world application, a potential device needs to work with unseen subjects.

\paragraph{\textbf{Metrics.}} Exercise recognition and form evaluation tasks are treated as classification problems and their results are presented in terms of top-1 recognition accuracy. Results of exercise repetition counting are provided in  Mean Absolute Error (MAE) and absolute error values presented in percentage of all samples |e| where |e| equals 0, 1, 2 and greater than 2.

\paragraph{\textbf{Experimental setup.}}

Data augmentation is applied to the input sequences following \cite{buslaev2020albumentations}, including random resizing, rotation, cropping, greyscale, sepia, pixelization, brightness, contrast, and motion blur. Input sequences are uniformly sub-sampled for training, validation and testing. Models are trained with a batch size $b_s = 8$, AdamW optimiser, cross-entropy loss function, and a learning rate $l_r = 0.01$ reduced by a factor of 0.5 in epochs 10 and 15. $P_{NET}$ for the repetition counting module is trained with $b_s = 128$ and with an additional $l_r$ reduction after the 50th epoch. We initialise the learning process with Kinetics-400~\cite{kay2017kinetics} pre-trained model weights and ImageNet~\cite{russakovsky2015imagenet} for $P_{NET}$ with a frozen feature extraction part and gradually unfreeze the layers every 10th epoch starting from the last layer. Weights are stored for best validation accuracy.

\subsection{Exercise Recognition}

For equal distribution of training classes, we observe the accurate performance of recognition by SlowFast \cite{feichtenhofer2019slowfast} with 96.73\% accuracy, followed by Swin3D \cite{liu2022video} with 98.18\% and with the best result overall equal to 98.55 \% from MViTv2 \cite{li2022mvitv2}. These results are summarised in Table \ref{tab:exe_recog_equal}. In LOOCV evaluation, we observe that Swin3D \cite{liu2022video} is the least accurate for each subject, and MViTv2 \cite{li2022mvitv2} performs best for 6 subjects. The most challenging subjects are III and IX, resulting in 81.92\% and 86.03\% respectively, leaving room for further improvement. LOOCV results are summarised in Table \ref{tab:ex_rec_cross_subj}.

\input{tables/ex_ev}
\input{tables/ar_cross_user}

\subsection{Exercise Form Evaluation}

In equally distributed data, we observe the poor performance of the Swin3D backbone with 63\% accuracy of exercise form evaluation, where training leads to vanishing gradients. MViTv2 is in the middle resulting in 83.26\%. The best results are obtained with the SlowFast equal to 86.98\%. 
The LOOCV scenario confirms the high performance of the SlowFast feature extractor. However, in this scenario the effect of vanished gradients is not observed for Swin3D, resulting in a performance close to SlowFast. Exercise form evaluation remains the most challenging for MViTv2. LOOCV evaluation increases the complexity, with the worst result for subject VII equal to 41.1\%. All results are summarised in Table~\ref{tab:ex_rec_cross_subj}. 

\subsection{Repetition Counting}

\input{tables/rep_count}

In the equally distributed test subset, our approach results in MAE of 1.33 with 66.1\% of samples correctly counted, 15\% of samples falling in the absolute error of 1 repetition |e| = 1, 6.1\% |e| =2 and 12.8\% |e| > 2. Per-exercise performance is analysed without distinguishing between the left or right hand, as the movement follows a similar pattern regardless of the hand used. The best results are obtained for exercises I-III, where subjects tend to perform repetitions more slowly due to the nature of the exercise. In exercises III, IV, V and VI, we observe the method to decrease performance as some of the subjects start the exercise from different hand positions, which is correct in the case of these exercises but creates a challenge for the algorithm. The lowest performance is observed in XII, where the movement is minimalist, thereby causing a challenge to detect the pick, and XIII, where the exercise is performed rapidly and it is appropriate to initiate the hand movement in any direction. All of the results are presented in Table \ref{tab:rep_count}. The performance drops significantly for the LOOCV assessment, which is depicted in Table \ref{tab:ex_rec_cross_subj}. In this case, pick detection is trained separately for all scenarios. For the best subject, the MAE is 3.29. The worst performance is 6.96.

\subsection{Inference times}
\begin{figure}[t]
  \centering
  \includegraphics[width=1\linewidth]{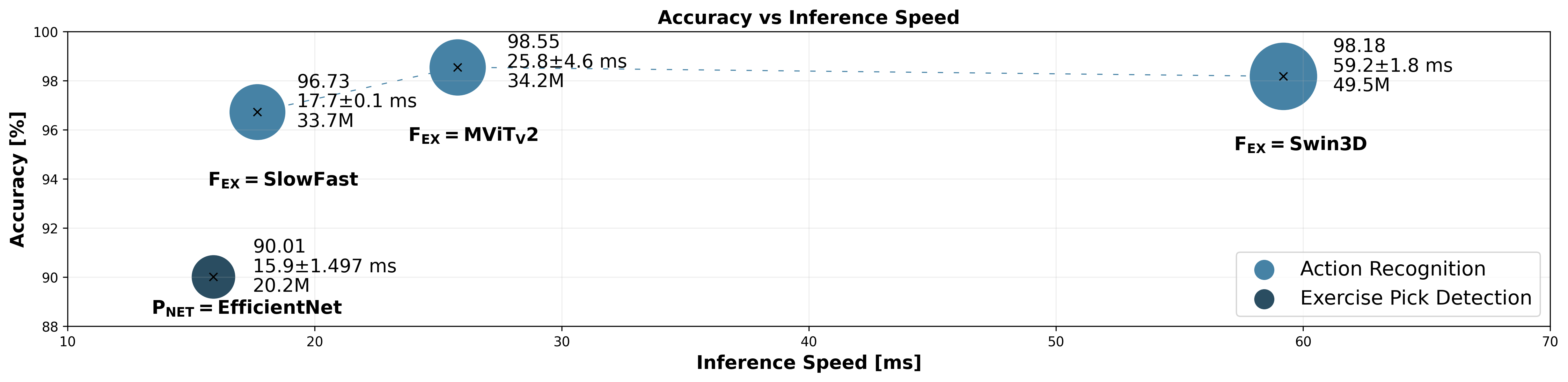}
  \caption{Inference times for exercise recognition and pick detection with their accuracy. Each method is visualised as a circle whose size represents the number of parameters.}
  \label{fig:inference}

\end{figure}

We measure the inference times by averaging 1000 trials on the NVIDIA GeForce RTX3090 GPU. First, we assess the performance for exercise recognition with different $F_{EX}$ components. For the exercise form evaluation, we observe similar times as the only difference in the architecture is the output dimension of the last layer of the prediction head, which is changed from $H_{REC}$ to $H_{FORM}$, reducing it from 25 to 2. The fastest inference is observed with SlowFast, followed by MViTv2. The slowest performance is with Swin3D, which is 3 times slower than SlowFast. Furthermore, it is not possible to measure the inference of the exercise repetition counter as it works continuously and considers different lengths of sequences. We measure the time of the exercise pick estimation with $P_{NET}$, which estimates the pick with 90.01\% accuracy in 15.9 ms.

\subsection{Ablation Studies}

\paragraph{\textbf{Initialisation strategy for exercise form evaluation.}}

We employ state-of-the-art video understanding networks with transfer learning through pre-trained model weights. This technique is confirmed to be powerful in tasks such as video classification but raises questions regarding the best strategy for the task of exercise form evaluation, which motivates us to experiment with different network initialisation strategies. The results are presented in Table \ref{tab:ablation}. In \textit{Ablation I} we perform training of the exercise form evaluation network with Kinetics-400 \cite{kay2017kinetics} weights frozen throughout the training process, which results in 77.21\% accuracy. By unfreezing the last layer of the backbone, the result increases to 86.05 (see \textit{Ablation II}). Furthermore, in \textit{Ablation III} we repeat the process with the weights derived during exercise recognition training and gain 80\% accuracy. The best results are observed by unfreezing the last layer in later epochs and initialising the architecture with weights pre-trained during exercise recognition (see Ablation IV). The experiment highlights the possibility of using common weights for both tasks with only a small performance penalty. This observation is important when aiming for real-time performance, as it reduces the computational cost by using only a single forward pass through the feature extraction part.

\input{tables/ablation}

\paragraph{\textbf{Filtering module for repetition counting.}}

The key component of repetition counting module $C$ is a filter $FIL$ built from noise spike filters $FIL_0$ and $FIL_1$.
We perform an ablation study using different $FIL$ setups to achieve the best performance by manipulating the value of spike noise length $N_{len}$ for $FIL_0$ and $FIL_1$ in the range from 1 to 6. The results are presented in Table \ref{tab:rep_count_abl}. We observe that filtering reduces MAE from 12.34 to 1.33. The best results are achieved by removing sequences from 1 to 5 with $FIL_1$ and 1 to 3 with $FIL_0$.

\input{tables/rep_count_abl}

%% file: tables/ex_ev.tex
\begin{table}[t]
  
  \label{tab:ablation}
  \centering
  \begin{minipage}{0.45\linewidth}
    \centering
    \caption{Exercise recognition in the subsets with equal distribution.}
     \begin{tabular*}{\textwidth}{l@{
\extracolsep{\fill}}cc}
    
      \toprule
       Method & Val. [\%] $\uparrow$& Test [\%] $\uparrow$\\
      \midrule
    \textit{SlowFast+}$H_{REC}$ & 97.45 & 96.73 \\
     \textit{Swin3D+}$H_{REC}$ & 98.18 & 98.18 \\
     \textit{MViTv2+}$H_{REC}$ & \textbf{99.64} & \textbf{98.55} \\
      \bottomrule
      \label{tab:exe_recog_equal}
    \end{tabular*}
  \end{minipage}
  \hspace{0.05\linewidth}
  \begin{minipage}{0.45\linewidth}
    \centering
    \caption{Exercise form evaluation in the subsets with equal distribution.}
    
    \begin{tabular*}{\textwidth}{l@{
\extracolsep{\fill}}cc}
      \toprule
      Method & Val. [\%] $\uparrow$& Test [\%] $\uparrow$\\
      \midrule
      \textit{SlowFast+}$H_{FORM}$ & \textbf{94.92} & \textbf{86.98} \\
       \textit{Swin3D+}$H_{FORM}$& 68.02 & 63.26 \\
       \textit{MViTv2+}$H_{FORM}$& 91.37 & 83.26 \\
      \bottomrule
      \label{tab:recognition}
    \end{tabular*}
  \end{minipage}

\end{table}

%% file: tables/ar_cross_user.tex
\begin{table}[t]
  \caption{The results of the LOOCV test for all three tasks for each subject S.
  }
  \label{tab:headings}
  \centering
  \begin{tabular*}{\textwidth}{l@{
\extracolsep{\fill}}ccccccccc}
    \toprule
    Method & S-I & S-II & S-III & S-IV & S-V & S-VI& S-VII & S-VIII & S-IX\\
    \midrule
   \multicolumn{10}{c}{Exercise recognition - accuracy [\%] $\uparrow$} \\
    \midrule
    \textit{SlowFast+}$H_{REC}$ & \textbf{96.50} & 93.31 & 69.00  & 85.53 & 86.45 & \textbf{86.60}  & 94.60 & \textbf{94.95}& 84.92 \\
    \textit{Swin3D+}$H_{REC}$  & 86.71 & 79.66 &55.35 & 82.36 & 79.55 &73.53 &90.15 &63.88 &72.06\\
    \textit{MViTv2+}$H_{REC}$ & 93.71 & \textbf{94.15} & \textbf{81.92} & \textbf{90.00} &\textbf{94.58} &85.57 &\textbf{97.46} &91.41 &\textbf{86.03} \\
   \midrule
   \multicolumn{10}{c}{Exercise form evaluation - accuracy [\%] $\uparrow$} \\
    \midrule
    \textit{SlowFast+}$H_{FORM}$ &\textbf{76.22} &83.84 &\textbf{82.53} &76.05 & 72.66 & 56.38 & \textbf{41.10} & 70.45 & \textbf{74.86}\\
    \textit{Swin3D+}$H_{FORM}$ &75.52 & \textbf{86.07} & 66.17 & 70.00 & \textbf{75.86} & \textbf{90.07} &37.86 & 62.63& 60.34\\
    \textit{MViTv2+}$H_{FORM}$ & 75.52 & 77.44 & 72.49 & \textbf{76.84} & 72.41 & 53.90 &38.19 & \textbf{71.72} & 69.27 \\
    \midrule
\multicolumn{10}{c}{Repetition counting - MAE $\downarrow$} \\
\midrule
$P_{NET}$\textit{+C}&4.13 & 4.92 &4.54 &6.38 &6.96 &4.46 &6.0 &6.62 &3.29 \\

  \bottomrule
  \label{tab:ex_rec_cross_subj}
  \end{tabular*}

\end{table}

%% file: tables/rep_count.tex
\begin{table}[tb]
  \caption{Repetition counting in the equally distributed test subset.}
  \label{tab:headings}
  \centering
  \begin{tabular*}{\textwidth}{l@{
\extracolsep{\fill}}ccccc}
    \toprule
Exercise type     & MAE $\downarrow$ & |e| = 0 [\%] $\uparrow$& |e| = 1 [\%]& |e| = 2 [\%]& |e| > 2 [\%]\\
    \midrule
I - Towel Hand Closing & \textbf{0.00} & \textbf{100.0} & \textbf{0.0} & \textbf{0.0} & \textbf{0.0}\\
  II - Finger Nerve Glide & \textbf{0.00} & \textbf{100.0} & \textbf{0.0} & \textbf{0.0} & \textbf{0.0}\\
  III - Wrist Curl & 0.04 & 96.0 & 4.0 & 0.0 & 0.0 \\
  IV - Hand Slide & 0.31 &76.9& 15.3& 7.6& 0.0\\
  V - Wrist Flexion and Ext. & 0.38 & 62.5 & 37.5 & 0.0 & 0.0\\
  VI - Pen Slide & 0.55 & 72.7 & 9.1 & 9.1 & 9.1 \\
  VII - Pen Spin & 0.61 & 61.1 & 16.6 & 22.2 & 0.0 \\ 
VIII - Coin Drop & 0.80 & 46.7 & 33.3 & 13.3 & 6.7\\ 
IX - Palm Up and Down & 1.00 & 60.0 & 20.0 & 0.0 & 20.0\\
X - Ball Grip & 1.00 & 66.6 & 6.6 & 6.6 & 20.0 \\
XI - Wrist Extension & 1.61 & 73.9 & 4.4 & 0.0 & 21.7\\
XII - Rolling Bottle & 2.70 & 43.3 & 30.0 & 6.6 & 20.0 \\
XIII - Pushing Hands & 10.67 & 0.0 & 0.0 & 0.0 & 100.0\\
 \midrule
    Overall & 1.33 & 66.1 & 15.0 & 6.1 & 12.8 \\
  \bottomrule
  \label{tab:rep_count}
  \end{tabular*}

\end{table}

%% file: tables/ablation.tex
\begin{table}[t]
  \caption{Exercise form evaluation according to the weight initialisation technique.}
  \label{tab:headings}
  \centering
  \begin{tabular*}{\textwidth}{l@{
\extracolsep{\fill}}cccc}
    \toprule
     & Weights & Unfreezed & Validation $\uparrow$& Test $\uparrow$\\
    \midrule
    Ablation I &Kinetics-400 & \ding{55} & 85.79 & 77.21  \\
    
    Ablation II &Kinetics-400 & \checkmark & 95.43 & 86.05 \\
    Ablation III & Ex. Recognition & \ding{55} & 87.82 & 80.00 \\
  Ablation IV & Ex. Recognition & \checkmark& \textbf{94.92} & \textbf{86.98} \\
  \bottomrule
  \label{tab:ablation}
  \end{tabular*}

\end{table}

%% file: tables/rep_count_abl.tex
\begin{table}[tb]
  \caption{Results of repetition counting with different $FIL$ configurations.}
  \label{tab:headings}
  \centering
  \begin{tabular*}{\textwidth}{l@{
\extracolsep{\fill}}ccccccccccccc}
    \toprule

    & \multicolumn{6}{c}{$FIL_1$} & \multicolumn{6}{c}{$FIL_0$} \\
\midrule    
$N_{len}$ &1 &2 &3 &4 &5 &6 &1 &2 &3 &4 &5 &6 &MAE $\downarrow$\\

\midrule
I & \ding{55} & \ding{55} & \ding{55} & \ding{55} & \ding{55} & \ding{55} & \ding{55} & \ding{55} & \ding{55} & \ding{55} & \ding{55} & \ding{55} &12.34\\

II & \checkmark & \ding{55} & \ding{55} & \ding{55} & \ding{55} & \ding{55} & \ding{55} & \ding{55} & \ding{55} & \ding{55} & \ding{55} & \ding{55} &5.77\\

III & \checkmark & \checkmark & \ding{55} & \ding{55} & \ding{55} & \ding{55} & \ding{55} & \ding{55} & \ding{55} & \ding{55} & \ding{55} & \ding{55} &3.42\\

IV & \checkmark & \checkmark & \checkmark & \ding{55} & \ding{55} & \ding{55} & \ding{55} & \ding{55} & \ding{55} & \ding{55} & \ding{55} & \ding{55} &2.29\\

V & \checkmark & \checkmark & \checkmark & \checkmark & \ding{55} & \ding{55} & \ding{55} & \ding{55} & \ding{55} & \ding{55} & \ding{55} & \ding{55} &1.86\\

VI & \checkmark & \checkmark & \checkmark & \checkmark & \checkmark & \ding{55} & \ding{55} & \ding{55} & \ding{55} & \ding{55} & \ding{55} & \ding{55} &1.6\\

VII & \checkmark & \checkmark & \checkmark & \checkmark & \checkmark & \checkmark & \ding{55} & \ding{55} & \ding{55} & \ding{55} & \ding{55} & \ding{55} &1.7\\

VIII & \checkmark & \checkmark & \checkmark & \checkmark & \checkmark & \ding{55} & \checkmark & \ding{55} & \ding{55} & \ding{55} & \ding{55} & \ding{55} &1.36\\

IX & \checkmark & \checkmark & \checkmark & \checkmark & \checkmark & \ding{55} & \checkmark & \checkmark & \ding{55} & \ding{55} & \ding{55} & \ding{55} &1.35\\

X & \checkmark & \checkmark & \checkmark & \checkmark & \checkmark & \ding{55} & \checkmark & \checkmark & \checkmark & \ding{55} & \ding{55} & \ding{55} &\textbf{1.33}\\

XI & \checkmark & \checkmark & \checkmark & \checkmark & \checkmark & \ding{55} & \checkmark & \checkmark & \checkmark & \checkmark & \ding{55} & \ding{55} &1.37\\

XII & \checkmark & \checkmark & \checkmark & \checkmark & \checkmark & \ding{55} & \checkmark & \checkmark & \checkmark & \checkmark & \checkmark & \ding{55} &1.39\\

XIII & \checkmark & \checkmark & \checkmark & \checkmark & \checkmark & \ding{55} & \checkmark & \checkmark & \checkmark & \checkmark & \checkmark & \checkmark &1.43\\

  \bottomrule
  \label{tab:rep_count_abl}
  \end{tabular*}

\end{table}

%% file: sections/5_conclusion.tex
In this study, we proposed remote hand rehabilitation through egocentric vision and wearable smartglasses for people who have suffered a stroke. We evaluated the approach by creating the first dataset of egocentric videos of people performing rehabilitation exercises called REST-HANDS. It includes labels for exercise recognition, exercise form evaluation and repetition counting. We implemented baseline methods for all three tasks using state-of-the-art video understanding neural networks. Experiments on test subsets with evenly distributed samples resulted in exercise recognition of 98.55\%, form evaluation of 86.98\%, and repetition counting MAE equal to 1.33. Repetition counting worked accurately except for exercises where the movement is minimalistic, and it is difficult to detect its pick, the subject performs the exercise fast, which is on the edge of correct execution, or the the form is correct despite the starting direction of movement. 
We found that the LOOCV test was more inaccurate in all the tasks, as the models had to generalise. Exercise recognition worked best based on MViTv2, achieving 81.92\% accuracy for the most challenging subject and 97.46\% in the best case. Exercise form evaluation performed the best building on SlowFast, and it ranged from 41.1\% to 82.53\%. Repetition counting dropped in performance with results ranging from 3.29 to 6.96 MAE. These observations indicate a high potential for the development of automated tools for hand rehabilitation using egocentric vision. Differences in performance for each task highlight the need for further research into methodologies, particularly in assessing exercise form and counting repetitions. REST-HANDS is a step forward in bringing automated remote hand rehabilitation to those in need, addressing the critical need for solutions to deal with an ageing society that requires more and more medical professionals.